# Visualizing Semantic Structures of Sequential Data by Learning Temporal Dependencies

**Kyoung-Woon On**
Department of Computer Science and Engineering
Seoul National University
Seoul, South Korea

**Eun-Sol Kim**
KaKao Brain
Seongnam, South Korea

**Yu-Jung Heo and Byoung-Tak Zhang**
Department of Computer Science and Engineering
Seoul National University
Seoul, South Korea

## Abstract

While conventional methods for sequential learning focus on interaction between consecutive inputs, we suggest a new method which captures composite semantic flows with variable-length dependencies. In addition, the semantic structures within given sequential data can be interpreted by visualizing temporal dependencies learned from the method. The proposed method, called Temporal Dependency Network (TDN), represents a video as a temporal graph whose node represents a frame of the video and whose edge represents the temporal dependency between two frames of a variable distance. The temporal dependency structure of semantic is discovered by learning parameterized kernels of graph convolutional methods. We evaluate the proposed method on the large-scale video dataset, Youtube-8M. By visualizing the temporal dependency structures as experimental results, we show that the suggested method can find the temporal dependency structures of video semantic.

## Introduction

One fundamental task in learning sequential data is to find temporal dependencies among successive frames and to learn representations of the sequential data using dependencies. Conventional methods for learning sequential data focus on temporal relationships between a few successive frames, because the model complexity grows along with the number of successive frames to be modeled.

In terms of neural network architectures, many problems with sequential inputs are resolved by using Recurrent Neural Networks (RNNs) and their variants as it naturally takes sequential inputs frame by frame. However, as RNN-based methods take frames in (incremental) order, the parameters of the methods are trained to capture patterns in transitions between successive frames, making it hard to find long-term temporal dependencies through overall frames. For this reason, their variants, such as long short-term memory (LSTM) (Hochreiter and Schmidhuber 1997) and gated recurrent units (GRU) (Chung et al. 2014), have made the suggestion of ignoring noisy (unnecessary) frames and maintaining the semantic flow with gating mechanism. However, it is still hard to retain multiple semantic flows and to interpret various dependencies between frames.

In this work, we propose a graph-based neural network architecture, which learns representations of sequential data while considering the temporal dependency structure of it. The sequential data is represented as a graph, in that a node represents each frame and the relationship between two frames is represented as an edge between two nodes. We define the temporal dependency structure through overall frames as a weighted adjacency matrix of the graph. The adjacency matrix is estimated by learning parameterized kernels of node representations. Furthermore, the representation of each node (frame) is learned by applying spectral graph theory with the adjacency matrix. The methods to find the temporal dependency structure and to learn representations of the sequential inputs can be learned in an end-to-end manner.

We evaluate our method with the YouTube-8M dataset, which contains about 6.1M videos and 3862 labels. As qualitative analysis of the proposed model, we visualize semantic temporal dependencies of sequential input frames which are automatically constructed and interpret these dependencies with a real video example.

The remainder of the paper is organized as follows. To make further discussion clear, mathematical definitions and notations of graphs are clarified. Subsequently, the problem statement of this paper are described in the following section. After that, the suggested method, called Temporal Dependency Network (TDN) is suggested in detail and the experimental results with the real dataset YouTube-8M is shown.

## Graph notations

A graph $\mathcal{G}$ is denoted as a pair $(V, E)$ with $V = \{v_1, ..., v_n\}$ the set of nodes (vertices), and $E \in V \times V$ the set of edges. Each node $v_i$ is associated with a feature vector $x_i \in \mathbb{R}^m$. To make notations compact, the feature matrix of graph $\mathcal{G}$ is denoted as $X = \{x_1^\top, x_2^\top, ..., x_N^\top\} \in \mathbb{R}^{N \times m}$. Also, a graph has an $N \times N$ weighted adjacency matrix $A$ where $A_{i,j}$ represents the weight of the edge between $v_i$ and $v_j$.

## Problem Statement

Sequential video data can be expressed in the form of a graph structure, which presents each frame of video as a node and the dependency between each pair of nodes as an

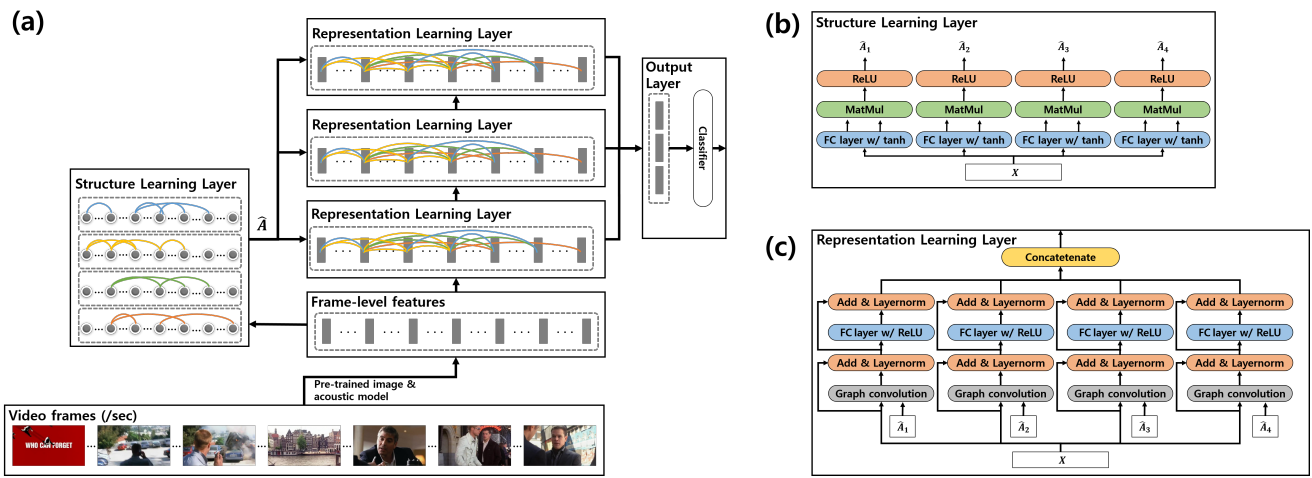

Figure 1: (a): Overall architecture of the Temporal Dependency Network (TDN) for a video classification task. (b), (c): sophisticated illustrations of Structure learning layer and Representation learning layer in the case of the number of dependency structure $N = 4$.

edge. More specifically, suppose that we have a video data which has $N$ successive frames $V = \{v_1, ..., v_n\}$. The representations of frames can be expressed in a matrix form: $X = \{x_1^T, x_2^T, ..., x_N^T\} \in \mathbb{R}^{N \times m}$, where $x_i \in \mathbb{R}^m$ denotes representations of $i$-th frame. Also, the dependency between two frames $v_i, v_j$ can be presented as an weighted edge $e_{ij} \in E$. Then, the dependency structure among video frames can be completely defined by the weighted adjacency matrix $A$, where $A_{ij} = e_{ij}$. With the aforementioned notations and definitions, we can now formally define the problem of sequential structure learning for video representations as follows:

*Given video frames representations $X \in \mathbb{R}^{N \times m}$, we seek to discover a weighted adjacency matrix $A \in \mathbb{R}^{N \times N}$ which represents dependency among frames.*

$$f : X \to A \quad (1)$$

With $X$ and $A$, new representations for video frames $H \in \mathbb{R}^{N \times m}$ are acquired by $g$.

$$g : \{X, A\} \to H \quad (2)$$

The obtained video representations $H$ can be used for various tasks of video understanding. In this paper, multi-label classification problem for the determination of the key topical themes of a video is mainly dealt with.

## Temporal Dependency Network

The suggested method, the Temporal Dependency Network (TDN), consists of two sub-modules: the structure learning layer and the representation learning layer. In the structure learning layer, the temporal dependency structure is constructed by estimating a weighted adjacency matrix $\hat{A}$ via parameterized kernels over $X$. With $\hat{A}$, the representations of each frame are trained based on a convolutional mechanism on graph in the representation learning layer. In the next sections, detail operations of each of these modules are described.

### Structure Learning Layer

To estimate the temporal dependency structure over all frames, the weighted adjacency matrix $\hat{A}$ is learned by the parameterized kernel $\mathcal{K}$:

$$\hat{A}_{ij} = \mathcal{K}(x_i, x_j) = ReLU(f(x_i)^\top f(x_j)) \quad (3)$$

where $f(x)$ is a single-layer feed-forward network:

$$f(x) = \tanh(W^f x + b^f) \quad (4)$$

with $W^f \in \mathbb{R}^{m \times m}$ and $b^f \in \mathbb{R}^m$.

To give $\hat{A}$ the property of non-negativity, various functions (e.g., $exp()$ function) can be used as a final activation. In the case of a large graph, however, a dense adjacency matrix followed by normalization generally has noisier information. To eliminate this problem, we drive the graph to get more sparse connections by simply using the rectified linear unit ($ReLU$).

### Representation Learning Layer

After estimating the weighted adjacency matrix $\hat{A}$, the representation learning layer updates the representations of each frame via a graph convolution operation (Kipf and Welling 2016) followed by a position-wise fully connected feed-forward network. We also employ a residual connection (He et al. 2016) around each layer followed by layer normalization (Ba, Kiros, and Hinton 2016):

$$Z = LN(\sigma(\hat{D}^{-\frac{1}{2}} \hat{A} \hat{D}^{-\frac{1}{2}} X W^Z) + X) \quad (5)$$

$$H = LN(\sigma(ZW^H + b^H) + Z) \quad (6)$$

where $W^Z, W^H \in \mathbb{R}^{m \times m}$ and $b^H \in \mathbb{R}^m$. The representation layer can be stacked multiple times to learn higher-level abstracted representations.

## Multiple Dependency Structure

Instead of a single weighted adjacency matrix $\hat{A}$, we can estimate multiple adjacency matrices to capture various dependencies of frames. Concretely, in order to get $K$ adjacency matrices, the structure learning layer simply has $K$ different non-linear projections $f_k$. In the same way, the representation learning layer inputs different $\hat{A}_k$ to differently parameterized modules, then concatenate each output $H_k$ to construct a final $H$. To prevent over-parameterization, we reduce the dimensions of each output, by setting $W^f, W^Z, W^H \in \mathbb{R}^{m \times (m/K)}$ and $b^f, b^H \in \mathbb{R}^{m/K}$. By doing so, the total computational cost is similar to that of a single adjacency matrix with full dimensionality. Figure 1 (b) and (c) illustrate detailed computations of the structure learning layer and the representation learning layer in the case of $K = 4$.

## Example: TDN for Video Classification

From there on, the suggested model is instantiated with a video classification problem. A large-scale video dataset is used as the sequential input to the TDN, and the problem task is to predict multi-labels when a video clip is given. The video inputs consist of two multimodal sequences, which are the image and acoustic features. Each feature of two modalites is extracted from pre-trained deep learning models (Abu-El-Haija et al. 2016), and the extracted features are concatenated to construct a frame-level feature vector $\mathbf{x} \in \mathbb{R}^m$. After applying structure and representation learning layer to the frame-level features, four different dependency structures among frames are extracted. Also, intermediate frame-level features from the each layer of three representation learning layers are concatenated, and then the final video-level feature is constructed by an average pooling operation over all of the frame-level features. Finally, multiple labels for a video-level feature are predicted using simple logistic regression in the classification module. Figure 1 illustrates the whole TDN structure.

# Experiments

## YouTube-8M Dataset

YouTube-8M (Abu-El-Haija et al. 2016) is a benchmark dataset for video understanding, where the main task is to determine the key topical themes of a video. It consists of 6.1M video clips collected from YouTube, and because of the large scale of the dataset visual/audio features are pre-extracted and provided with the dataset. Each video is labeled with one or multiple tags referring to the main topic of the video and it is extracted from Knowledge Graph entities. The videos are split into 3 partitions, which are 70% for training, 20% for validation and 10% for test. As we have no access to the test labels, all results in this paper are reported for our validation set.

Each video is decoded at 1 frame-per-second up to the first 360 seconds. More specifically, the frame-level visual/audio features were extracted by using a state-of-the-art deep learning model: inception-v3 network (Szegedy et al. 2016) trained on imagenet and VGG-inspired architecture (Hershey et al. 2017) trained for audio classification are used for image and audio feature respectively. PCA and whitening are then applied to reduce the dimensions to 1024 for the visual features and 128 for the audio features.

## Experimental Results: Visualizing Semantic Structures

The various types of semantic flow constitute scenes and events in the video and is constituted by multi-level and multi-scale composition of frames. In this section, we demonstrate capability of the proposed model to learn various temporal dependency structures to capture these semantic flows. Figure 2 shows a real example to analyze learned dependency structures.

At the top of the figure, four dependency structures are represented by (symmetrically normalized) weighted adjacency matrices, coded in a gray-scale image. Along with diagonal elements of the adjacency matrices, bright blocks can be found which present highly connected frames. We found that those groups of frames (the diagonal blocks) constitute semantic events in various lengths (Figure 2 (b)). For example, the event #7 in the dependency structure 1 indicates the scene of multiple players in tactical training. the event #9 and the #11 in the dependency structure 1 and 3 are the scene that a player steps off the court and gets coached during the training session and the scene that the coach and the players are interviewed, respectively. The events from #1 to #14 (color-coded by blue) are other examples which can be seen as semantically plausible event scenes.

Another interesting result is founded in off-diagonal bright blocks. The off-diagonal bright blocks show that the model also capture the dependencies between events. For examples, the event #15, #16 and #17 reveals that the event #4, #6 and latter part of the #2 are highly correlated. Event #4 and the latter part of the #2 are the scene of players inside the therapy room being treated by medical staff. Interestingly, event #6 is the scene that players help each other stretch on the basketball court. The events from #15 to #21 (color-coded by red) are other examples which shows the dependencies between events (in the diagonal blocks).

The detailed description of each events are as follows: #1: Scene of the external view of a basketball stadium. #2:The external view of the stadium is shown and the scenes changes immediately to players inside a therapy room being treated by medical staff. #3: Some players are warming up on the court. #4: scene of players inside the therapy room being treated by medical staff. #5: A view of the court as the warm-up session is wrapping up. #6: Players help each other stretch on the basketball court. #7: Multiple players in tactical training. #8: Scene that shows training only on the basketball court. #9: A player steps off the court and gets coached during the training session. #10: A collection of scenes on court, including a player being coached, an interview and other multiple coaching scenes. #11: The coach and the players are interviewed. #12: While a player is being interviewed, the players get coached individually. #13: the players are coached as a team. #14:(Entire frames) A behind-the-scenes look of basketball players and their coach as they prepare for their upcoming games. #15-#17: corre-lations among the events latter part of #2, #4 and the #6. #18-#20: correlations among the events #3, #5 and the #8. #21:the correlations between the former part of the #10 and the #11.

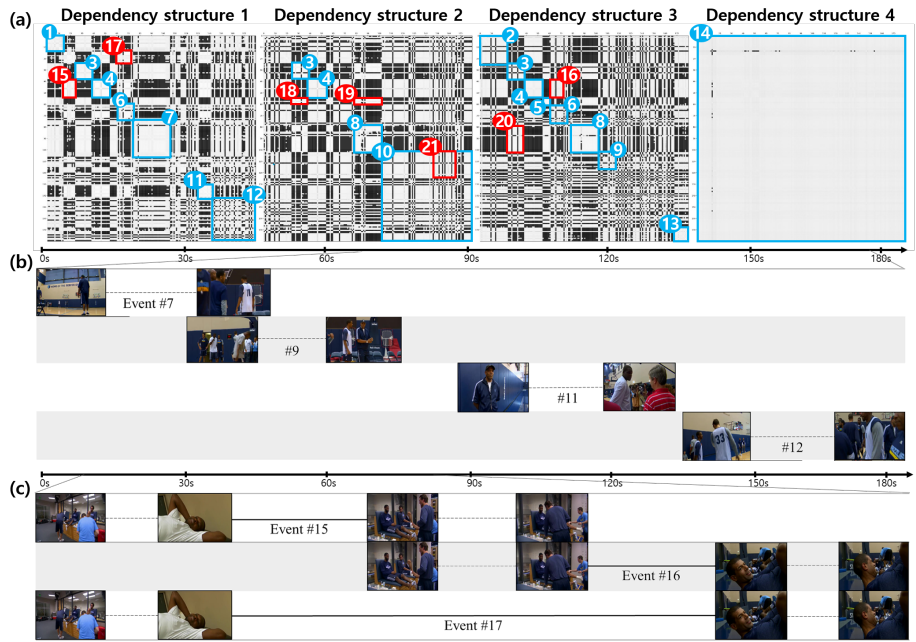

Figure 2: An example of the constructed temporal dependency structures for a real input video https://youtu.be/3GsHimDtPTI is visualized. (a) Four temporal dependency structures are represented by weighted adjacency matrices $\hat{A}$. The size of the matrix is equal to the length of the input video. The strength of connection between two frames are encoded in a gray-scale which 1 to white and 0 to black. In (b), 4 semantically interesting events of along with diagonal are shown. The diagonal bright blocks stands for consecutive frames and has various length. In (c), 3 events from off-diagonal blocks are represented. Those events can represent more complex semantic flows as those connects various events represented in diagonal blocks.

## Conclusions

In this paper, we proposed Temporal Dependency Networks (TDNs) which learn not only the representations of a sequence but also various temporal dependency structures within the sequence. The experiment is conducted on a real large-scale video dataset and it shows that the proposed model efficiently learns the inherent dependency structure of temporal data and discovers better representations.

## Acknowledgements

This work was partly supported by the Institute for Information & Communications Technology Promotion (2015-0-00310, 2017-0-01772, 2018-0-00622) and Korea Evaluation Institute of Industrial Technology (10060086) grant funded by the Korea government (MSIP, DAPA).